\documentclass[sigconf]{acmart}
\AtBeginDocument{%
  }

\renewcommand\footnotetextcopyrightpermission[1]{}
\setcopyright{acmlicensed}
\copyrightyear{2026}
\acmYear{2026}
\acmDOI{XXXXXXX.XXXXXXX}
\acmConference[Under Review]{Submission to ACM Conference}{2026}{Shanghai, China}
\acmISBN{}

\usepackage{multirow}

\begin{document}

\title{Hierarchical Memory Orchestration for Personalized Persistent Agents}





\author{Junming Liu}
\email{liujunming@pjlab.org.cn}
\affiliation{%
  \institution{Shanghai Artificial Intelligence Laboratory}
  \city{Shanghai}
  \country{China}
}

\author{Yifei Sun}
\email{sunyifei@pjlab.org.cn}
\affiliation{%
  \institution{Shanghai Artificial Intelligence Laboratory}
  \city{Shanghai}
  \country{China}
}

\author{Weihua Cheng}
\email{chengweihua@pjlab.org.cn}
\affiliation{%
  \institution{Shanghai Artificial Intelligence Laboratory}
  \city{Shanghai}
  \country{China}
}

\author{Haodong Lei}
\email{leihaodong@pjlab.org.cn}
\affiliation{%
  \institution{Shanghai Artificial Intelligence Laboratory}
  \city{Shanghai}
  \country{China}
}

\author{Yuqi Li}
\email{yuqili010602@gmail.com}
\affiliation{%
  \institution{The City University of New York}
  \city{New York}
  \country{USA}
}

\author{Yirong Chen}
\email{chenyirong@pjlab.org.cn}
\affiliation{%
  \institution{Shanghai Artificial Intelligence Laboratory}
  \city{Shanghai}
  \country{China}
}

\author{Ding Wang}
\authornote{Corresponding author.}
\email{wangding@pjlab.org.cn}
\affiliation{%
  \institution{Shanghai Artificial Intelligence Laboratory}
  \city{Shanghai}
  \country{China}
}

\renewcommand{\shortauthors}{Liu et al.}

\begin{abstract}
    While long-term memory is essential for intelligent agents to maintain consistent historical awareness, the accumulation of extensive interaction data often leads to performance bottlenecks.
    Naive storage expansion increases retrieval noise and computational latency, overwhelming the reasoning capacity of models deployed on constrained personal devices.
    To address this, we propose \textbf{H}ierarchical \textbf{M}emory \textbf{O}rchestration (\textbf{HMO}), a framework that organizes interaction history into a three-tiered directory driven by user-centric contextual relevance.
    Our system maintains a compact primary cache, coupling recent and pivotal memories with an evolving user profile to ensure agent reasoning remains aligned with individual behavioral traits.
    This primary cache is complemented by a high-priority secondary layer, both of which are managed within a global archive of the full interaction history.
    Crucially, the user persona dictates memory redistribution across this hierarchy, promoting records mapped to long-term patterns toward more active tiers while relegating less relevant information.
    This targeted orchestration surfaces historical knowledge precisely when needed while maintaining a lean and efficient active search space.
    Evaluations on multiple benchmarks achieve state-of-the-art performance.
    Real-world deployments in ecosystems like OpenClaw demonstrate that HMO significantly enhances agent fluidity and personalization.
\end{abstract}


\begin{teaserfigure}
  \fbox{\includegraphics[width=0.98\textwidth]{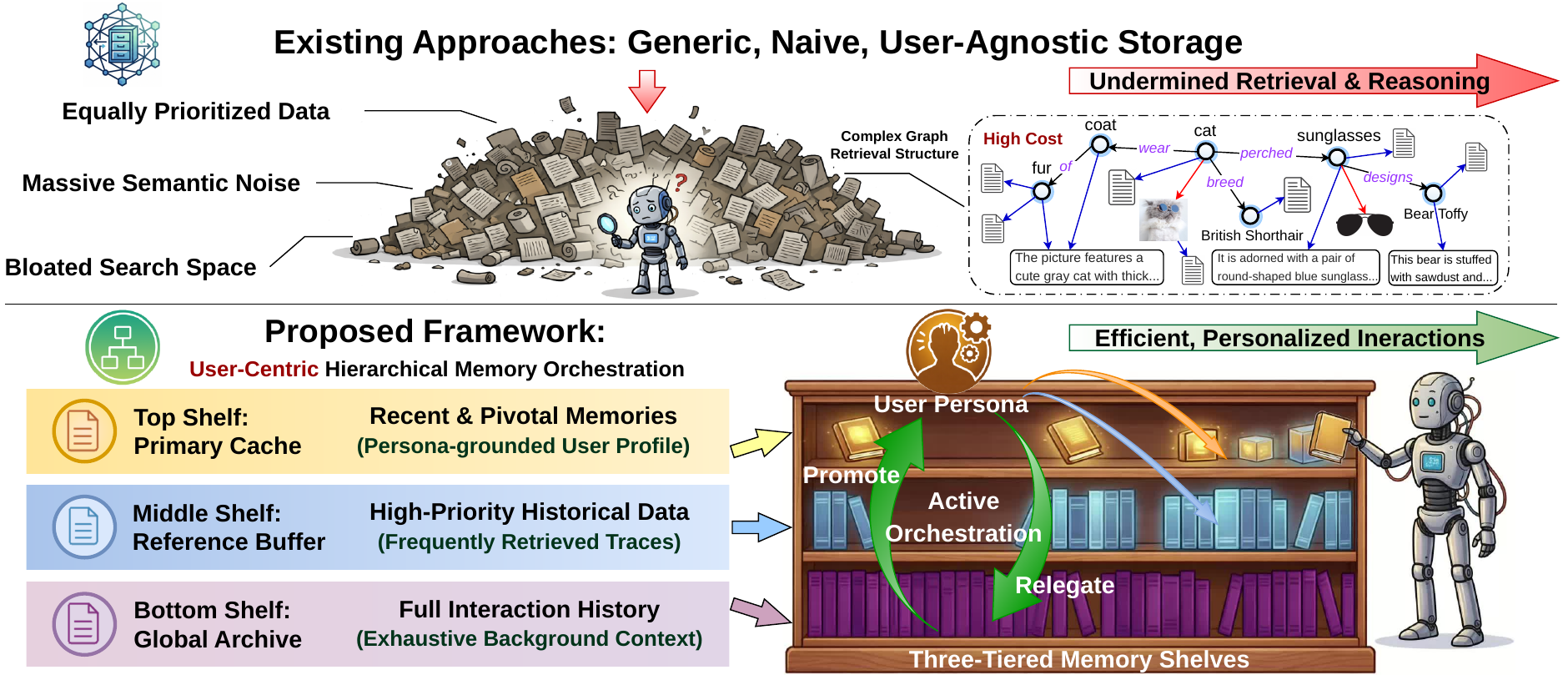}}
  \caption{Comparison between (top) existing user-agnostic storage and (bottom) our proposed method driven by user persona.}
  \Description{Comparison between (top) existing user-agnostic storage and (bottom) our proposed method driven by user persona.}
  \label{fig:teaser}
\end{teaserfigure}


\maketitle

\section{Introduction}

The rapid evolution of Large Language Models (LLMs) \cite{Grattafiori_2024_llama3, Yang_2025_Qwen3, Anthropic_2024_Claude3} and Multimodal Large Language Models (MLLMs) \cite{Wang_2025_Internvl3.5, Team_2026_Kimi} has significantly advanced the capabilities of intelligent agents, enabling more natural and versatile digital assistants.
However, the fundamental quality of user interaction is often compromised by the inability of current agents to recall past experiences reliably \cite{Wang_2023_Enabling, Packer_2024_MemGPT}. When an agent loses the thread of a long-term conversation, the resulting cognitive dissonance shatters the user's sense of immersion, leading to a fragmented experience that undermines the perceived intelligence and utility of the entire system \cite{Jones_2025_Users}.

While the necessity of memory is clear, scaling it presents a formidable technical challenge. As interaction history expands, the system must present relevant knowledge to the model by retrieving specific ground-truth fragments from a massive archival pool. The efficacy of this process is bounded by three interrelated challenges: \textit{memory storage efficiency}, \textit{retrieval precision}, and \textit{downstream inference capacity}. 
Existing research primarily concentrates on the latter two dimensions, employing sophisticated retrieval mechanisms \cite{Wang_2025_Mirix, Du_2025_MemR3, Liu_2025_Memverse} or enhancing reasoning performance through Reinforcement Learning from Human Feedback (RLHF) \cite{Christiano_2017_RLHF, Liu_2026_Hit-RAG} and Chain-of-Thought (CoT) prompting \cite{Wei_2022_CoT, Jin_2025_Disentangling}. However, these methods often require prohibitive computational costs for model fine-tuning or introduce substantial inference latency. Such delays are particularly detrimental in real-time user interaction, where prolonged response times undermine the fluid nature of the dialogue.
Furthermore, localized deployment restricts agents to smaller architectures \cite{Zheng_2025_A}. Executing complex memory workflows in these environments degrades performance and compromises user experience. This discrepancy between archival scale and local processing capacity renders traditional memory approaches ineffective.

Such practical constraints suggest that simply improving retrieval performance or scaling reasoning depth is insufficient, as these methodologies largely overlook the organization and personalization of the memory pool. By treating the storage layer as a flat library of linear document chunks or static graph nodes \cite{Chhikara_2025_Mem0, Rasmussen_2025_Zep, Liu_2025_Memverse, Li_2025_MemOS}, existing research implicitly adopts a strategy of naive expansion where every dialogue segment receives equal priority, as illustrated in Figure~\ref{fig:teaser}.
Crucially, these generic paradigms fail to account for individual variability in information utility across shifting user identities and contexts. They apply a rigid management logic that ignores how specific scenarios dictate the activation priority of past experiences. This lack of individualized orchestration means that memories essential to behavioral traits are often buried under generic noise. Consequently, the model must navigate a bloated search space, which inherently increases retrieval latency and undermines efficiency. Unlike complex graph-based structures that incur high computational costs during traversal, an adaptive, hierarchical approach to memory organization is essential to maintain the fluid nature of real-time interaction.

To reduce reliance on costly inference through architectural efficiency, we propose \textbf{H}ierarchical \textbf{M}emory \textbf{O}rchestration (\textbf{HMO}), a framework that transforms raw interactions into a structured multi-tiered memory system where data placement is governed by an evolving understanding of the user. Unlike static retrieval systems, HMO aligns immediate context with long-term personalization to support effective agent reasoning. This is achieved through a three-tiered hierarchy where each layer serves a distinct functional role. First, the primary cache maintains a fixed size, combining recent context with pivotal memories most relevant to the user. Then, an expandable secondary tier designed for high-priority historical data stores frequently retrieved traces, while a global archive persists the full interaction history. Crucially, the user persona guides the memory organization logic, promoting records aligned with individual traits to active layers while relegating less characteristic data to deeper storage.

During retrieval, HMO queries these tiers sequentially, retrieves knowledge through efficient mapping within the file system, and maintains a lean search space without redundant data movement.
Evaluations across multiple benchmarks and real-world deployment on the OpenClaw platform demonstrate that our framework achieves state-of-the-art performance and significantly enhances the coherence of the agent experience. Our main contributions are summarized as follows:
\begin{itemize}
    \item We are the first to employ an evolving user persona to uniquely dictate the entire memory lifecycle from prioritization through storage to retrieval.
    \item We propose a lightweight and plug-and-play framework that orchestrates a hierarchy of three tiers for efficient memory calling to ensure principled reasoning.
    \item Extensive evaluations and OpenClaw deployment demonstrate that HMO achieves state-of-the-art performance while ensuring highly personalized interactions for every user.
\end{itemize}

\section{Related Work}

\subsection{Memory-Augmented LLM Agents}
Memory enables (M)LLM-based agents to maintain long-term consistency in complex tasks \cite{Park_2023_Generative}. Early research focused on extending context via architectural modifications or external storage \cite{Zulfikar_2024_Memoro, Qian_2025_MemoRAG}, including MemGPT's paging mechanism \cite{Packer_2024_MemGPT} and Mem0's persistent layer for entity tracking \cite{Chhikara_2025_Mem0}. While effective, these systems primarily view memory as a flat, expanding repository, facing trade-offs between retrieval precision and computational efficiency as the search space grows.

\subsection{Architectural Memory Management}
Recent management strategies rely on flat vector databases or graph structures to store dialogue relations \cite{Wu_2025_A-MEM, Rasmussen_2025_Zep, Zhong_2024_MemoryBank}, often leading to bloated search spaces and high retrieval latency. To mitigate this, some frameworks utilize explicit deletion at the risk of data loss \cite{kang_2025_memory}, while parametric methods incur high fine-tuning costs \cite{Cao_2025_Memory, Liu_2025_Memverse}. Diverging from these, our multi-tiered architecture organizes history by utility to align storage with the user persona, enabling instant promotion while avoiding data loss and retraining.

\begin{figure*}[t]
    \centering
    \includegraphics[width=0.95\textwidth]{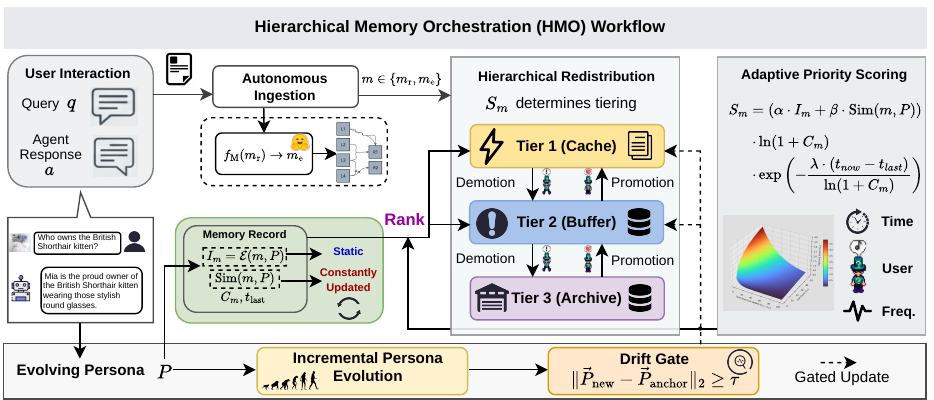}
    \caption{The system architecture and operational workflow of HMO.}
    \label{workflow}
\end{figure*}

\section{Methods}

HMO formalizes the memory lifecycle through four phases: autonomous ingestion, hierarchical redistribution, adaptive scoring, and incremental evolution, as illustrated in Figure~\ref{workflow}.

\subsection{Autonomous Memory Ingestion}

Upon generating a user query $q$ and the corresponding agent response $a$, the system captures the interaction as a discrete memory segment:
\begin{equation}
    m_{\text{r}} = (q, a).
\end{equation}
To ensure interaction fluidity, HMO supports preserving raw dialogue traces alongside extracted memory representations. For complex content such as extensive code snippets or technical documentation, the system employs a compression mapping:
\begin{equation}
    m_{\text{e}} = f_{\text{M}}(m_{\text{r}}),
\end{equation}
where $f_{\text{M}}$ denotes a MLLM that compresses interactions into specialized cognitive representations, such as semantic, episodic, or core memories \cite{Packer_2024_MemGPT, Wang_2025_Mirix, Liu_2025_Memverse}.
While compression risking information loss is a practical necessity for expansive contexts, it remains suboptimal for standard daily interactions. For shorter dialogues, HMO prioritizes the storage of $m_{\text{r}}$, thereby bypassing the information decay caused by conventional pruning and ensuring fidelity in subsequent recall.

Following capture, the agent executes an autonomous indexing process to evaluate the memory segment across multiple contextual dimensions. Specifically, an MLLM-based evaluator $\mathcal{E}$ analyzes the record $m$ (where $m \in \{m_{\text{r}}, m_{\text{e}}\}$) by considering the current user persona $P$ and the anticipated utility of $m$ relative to the user's immediate and future intent. This process yields an \textbf{Initial Importance} score:
\begin{equation}
    I_m = \mathcal{E}(m, P), \quad I_m \in [1, 10].
\end{equation}
The detailed scoring criteria are provided in Appendix A. We distinguish $I_m$ from the continuous semantic alignment $\text{Sim}(m, P)$, as empirical tests indicate that initialization based only on similarity is limited because an MLLM better accounts for more dimensions during initial scoring.
Although dynamic metadata updates could also be delegated to an MLLM, this approach prolongs computation and requires an elaborate design of rules. Since scoring is highly sensitive to these rules, any flaw in the design causes the results to deviate over time.
Consequently, we utilize $\mathcal{E}$ for initialization while relying on cosine similarity for efficient, real-time updates. Each record $m$ is then initialized with a header $\mathcal{H}_m$:
\begin{equation}
    \mathcal{H}_m = \{I_m, \text{Sim}(m, P), C_m, t_{\text{last}}\}.
\end{equation}

\subsection{Hierarchical Memory Redistribution}

Records are redistributed across three logical tiers based on their $S_m$ to maintain a lean active search space:
\begin{itemize}
    \item \textbf{Tier 1 (Cache):} This primary storage maintains $S$ recent conversation sessions alongside the top-$K$ pivotal memories. Any record accessed in this tier triggers a state refresh where $C_m$ increases and $t_{last}$ updates to $t_{now}$.
    \item \textbf{Tier 2 (Buffer):} This tier acts as a larger storage for $H$ records with high $S_m$ that do not fit into Tier 1. It functions as a secondary cache to intercept retrieval requests before they reach the global repository. Any record used in Tier 2 also triggers a metadata refresh and a potential promotion.
    \item \textbf{Tier 3 (Archive):} This tier serves as the global storage for all historical interactions. We only access this layer when the required information is missing from the active tiers. This hierarchical structure ensures that most queries are satisfied within the first two layers to minimize latency.
\end{itemize}

To ensure precision, the LLM performs a \textbf{self-reflective assessment} \cite{Asai_2024_Self-RAG} on Tier 1 context. If $q$ cannot be satisfied, it emits a trigger token to initiate a recursive search across Tier 2 and Tier 3. A successful retrieval increments $C_m$ and updates $t_{last}$, thereby increasing the record's $S_m$ for subsequent reasoning cycles.

\subsection{Adaptive Priority Scoring}

To maintain the accessibility of high-frequency records during idle periods, we implement a non-linear scoring function $S_m$ that incorporates an adaptive persistence mechanism:
\begin{equation}
\small
S_m = (\alpha \cdot I_m + \beta \cdot \text{Sim}(m, P)) \cdot \ln(1 + C_m) \cdot \exp\left(-\frac{\lambda \cdot (t_{now} - t_{last})}{\ln(1 + C_m)}\right).
\end{equation}
The scoring logic integrates $I_m$ and the dynamic persona alignment $\text{Sim}(m, P)$, weighted by coefficients $\alpha$ and $\beta$. The term $\ln(1 + C_m)$ provides a logarithmic gain based on the cumulative recall count $C_m$, while simultaneously acting as a resistance factor in the exponential denominator to modulate the base decay constant $\lambda$. As interaction frequency increases, the effective decay slope flattens, ensuring that memories with high utility $C_m$ remain prioritized in active tiers despite the elapsed time $t_{now} - t_{last}$.

To maximize efficiency, we restrict routine scoring updates to Tier 1 and Tier 2. Since $S_m$ governs the hierarchical distribution rather than the retrieval process, global calculation for the massive Tier 3 is unnecessary. During retrieval, we search through the tiers sequentially based on the query $q$, relying on similarity regardless of the score status. This ensures that records with high utility are retrieved even if their $S_m$ is not current. We adopt a lazy update strategy where a Tier 3 record is scored only when it is accessed. Upon retrieval, its $S_m$ is updated and ranked alongside all records in the active tiers. This leads to its promotion, demotion, or retention. While some records with high utility may temporarily stay in the bottom tier, this delay is acceptable. If a record in Tier 3 is not accessed, the existing context in the active tiers is already sufficient to handle $q$. This avoids the prohibitive cost of exhaustive computation and keeps system resources focused on active context management.

\subsection{Incremental Persona Evolution}

The persona $P$ is a dynamic entity that undergoes continuous updates. In principle, any shift in $P$ requires a full update of priority scores for all records in Tier 1 and Tier 2. This process specifically updates $\text{Sim}(\cdot, P)$ rather than $I_m$. Although a changing $P$ might eventually deviate from the initial logic of $I_m$, the retrieval algorithm adapts through the dynamic $Sim$ component. We omit re-scoring $I_m$ because invoking a MLLM for every update is not efficient. 

While general capacities for Tier 1 and Tier 2 are manageable, a user may choose to significantly increase these sizes to capture more active context. In such cases, the update costs would scale poorly with the larger memory pool. To maintain efficiency for these expanded active sets, we introduce a \textbf{Drift Gate} mechanism using an anchor persona $P_{\text{anchor}}$. A global update of scores and memory positions for the active tiers is triggered only when the deviation from $P_{\text{anchor}}$ exceeds a predefined threshold $\tau$:
\begin{equation}
\text{Update } \text{Sim}(\cdot, P) \text{ and set } P_{\text{anchor}} \leftarrow P_{\text{new}} \quad \text{iff} \quad \| \vec{P}_{\text{new}} - \vec{P}_{\text{anchor}} \|_2 \geq \tau.
\end{equation}
By benchmarking the current state $P_{\text{new}}$ against the last state that initiated a scoring cycle, we decouple high frequency persona refinement from expensive background sorting. This ensures that the memory distribution remains aligned with the user profile without exhaustive computation at every step.

\begin{table}[t]
\centering
\caption{Comparison of long-term memory performance on the LongMemEval dataset. Our proposed \textbf{HMO} outperforms existing RAG-based and memory-augmented agents in both retrieval recall and reasoning accuracy.}
\label{tab:longmemeval_hmo}
\setlength{\tabcolsep}{1pt}
\begin{tabular}{llccc}
\hline
\textbf{Method} & \textbf{Retriever} & \textbf{Rec@5} $\uparrow$ & \textbf{NDCG@5} $\uparrow$ & \textbf{Acc.} $\uparrow$ \\ \hline
No History \cite{Tan_2025_RMM} & - & - & - & 0.0 \\
Long Context \cite{Tan_2025_RMM} & - & - & - & 57.4 \\ \hline
RAG \cite{Tan_2025_RMM} & Stella & 59.2 & - & 61.4 \\
RAG \cite{Tan_2025_RMM} & GTE & 62.4 & - & 63.6 \\ \hline
MemoryBank \cite{Tan_2025_RMM} & Specific1 & 58.6 & - & 59.6 \\
LD-Agent \cite{Tan_2025_RMM} & Specific2 & 56.8 & - & 59.2 \\ 
A-Mem \cite{Li_2026_TiMem} & Qwen3-0.6B & - & - & 55.4 \\
Mem0 \cite{Li_2026_TiMem} & Qwen3-0.6B & - & - & 65.0 \\
MemoryOS \cite{Li_2026_TiMem} & Qwen3-0.6B & - & - & 58.0 \\
MemOS \cite{Li_2026_TiMem} & Qwen3-0.6B & - & - & 68.7 \\
RMM \cite{Tan_2025_RMM} & GTE & 69.8 & - & 70.4 \\
QRRetriever \cite{Zhang_2025_QRRetriever} & Llama-3.1-70B & 80.4 & - & 66.7 \\ \hline
\textbf{HMO (Ours)} & TE-small & \textbf{81.1} & \textbf{85.6} & \textbf{86.4} \\ \hline
\end{tabular}
\end{table}

\section{Experiment}

\subsection{Setups}

\textbf{Evaluation Datasets.}
To verify the effectiveness of HMO, we conduct experiments on two representative benchmarks for long-term conversational memory: (1) \textbf{LoCoMo} \cite{Maharana_2024_locomo}, which comprises multi-session dialogues across 10 distinct user groups, and (2) \textbf{LongMemEval-S} \cite{Wu_2025_LongMemEval}, a large-scale evaluation suite containing 500 conversations specifically tailored for assessing memory processing capabilities over extended horizons.

\textbf{Baselines.} We compare \textbf{HMO} with nine representative memory and retrieval frameworks: \textbf{MemoryBank} \cite{Zhong_2024_MemoryBank}, \textbf{Mem0} \cite{Chhikara_2025_Mem0}, \textbf{A-MEM} \cite{A-MEM}, \textbf{MemoryOS} \cite{Kang_2025_MemoryOS}, \textbf{MemOS} \cite{Li_2025_MemOS}, \textbf{LD-Agent} \cite{Li_2025_LD-Agent}, \textbf{RMM} \cite{Tan_2025_RMM}, \textbf{QRRetriever} \cite{Zhang_2025_QRRetriever}, and \textbf{MemVerse} \cite{Liu_2025_Memverse}. All models are evaluated using their recommended configurations.

\begin{sloppypar}
\textbf{Implementation Details.} 
We implement HMO using text-embedding-small for vectorization and GPT-4o-mini for response generation. To accommodate personalized interactions, we maintain a single memory stream for each conversation in LongMemEval-S, while keeping distinct memory profiles for different conversational roles in LoCoMo.
The hyperparameters are configured as $\alpha = 1$, $\beta = 1$, and $\tau = 0.10$. For the hierarchical storage, we set the primary tier size to $S = 5$ and $K = 50$, with the secondary storage capacity defined as $H = 200$.
\end{sloppypar}

\begin{figure*}[!t]
\centering
\fbox{\includegraphics[width=1.00\textwidth]{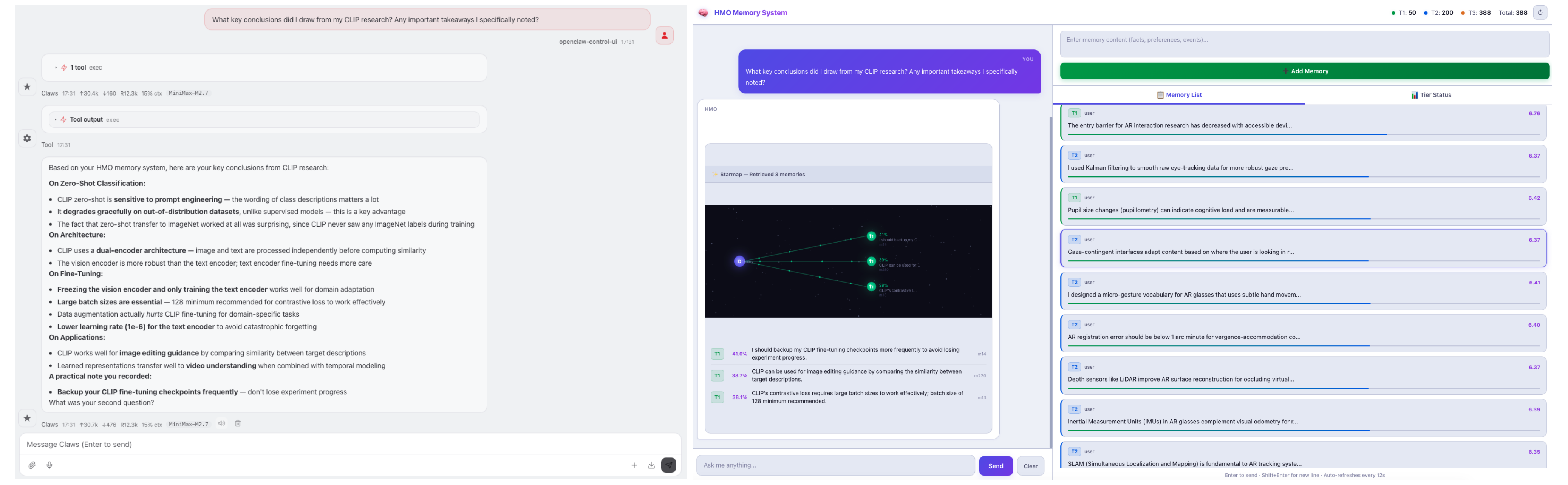}}
\caption{Demonstration of HMO application on OpenClaw, with a simple and informative user interface.}
\label{fig:openclaw_results}
\end{figure*}

\textbf{Evaluation Metrics.} 
For LoCoMo, we employ the F1 score to evaluate retrieval accuracy. For LongMemEval-S, we utilize Recall@5, NDCG@5, and LLM-as-a-judge to assess memory recall and generation quality. Specifically, Recall@5 is 1 only if all ground-truth documents are retrieved within the top-5, and 0 otherwise.

\subsection{Results}

\begin{table}[t]
\centering
\caption{Overall F1 performance on the LoCoMo benchmark using the GPT-4o-mini model.
}
\label{table:locomo_simple}
\setlength{\tabcolsep}{6mm}
\renewcommand{\arraystretch}{0.9}
\begin{tabular}{lc}
\toprule
\textbf{Method} & \textbf{Overall F1} $\uparrow$ \\ \midrule
A-Mem \cite{Liu_2025_Memverse} & 23.30 \\
MemoryBank \cite{Liu_2025_Memverse} & 31.42 \\
Mem0 \cite{Liu_2025_Memverse} & 42.13 \\
MemoryOS \cite{Liu_2025_Memverse} & 42.84 \\ 
MemVerse \cite{Liu_2025_Memverse} & 43.44 \\ \midrule
\textbf{HMO (Ours)} & \textbf{45.65} \\ \bottomrule
\end{tabular}
\end{table}

\textbf{Results on LongMemEval.} As shown in Table~\ref{tab:longmemeval_hmo}, HMO achieves state-of-the-art performance with 81.1\% Rec@5 and 86.4\% Acc. Using only the lightweight text-embedding-small, HMO outperforms QRRetriever based on Llama-3.1-70B at 80.4\% Rec@5. A high NDCG@5 of 85.6\% ensures that ground-truth memories are prioritized at the top of the context, leading to a peak accuracy of 86.4\%, which significantly exceeds the 66.7\% of the 70B-based baseline.

\textbf{Results on LoCoMo.} As reported in Table~\ref{table:locomo_simple}, HMO reaches an Overall F1 of 45.65 under the GPT-4o-mini backbone. It outperforms MemVerse at 43.44 and earlier memory-augmented agents such as MemoryBank at 31.42. These results confirm the robust generalizability of our hierarchical orchestration in maintaining high-quality memory profiles.
Detailed per-category metrics are provided in Appendix B.

\textbf{System Deployment and Evaluation.} We integrated the HMO memory system into AI agent platforms, including OpenClaw and Claude Code. As shown in Figure~\ref{fig:openclaw_results}, evaluations on OpenClaw demonstrate that HMO significantly reduces reasoning latency and accelerates task completion, confirming its efficiency in real-world agentic workflows.

\subsection{Ablation Study}

\begin{table}[t]
\centering
\renewcommand{\arraystretch}{0.9}
\setlength{\tabcolsep}{1mm}
\caption{Ablation results on LongMemEval.
}
\label{tab:ablation_hmo}
\begin{tabular}{l|cccc}
\toprule
\textbf{Configuration} & \textbf{Rec@5} $\uparrow$ & \textbf{NDCG@5} $\uparrow$ & \textbf{Acc.} $\uparrow$ & \textbf{Time (m)} $\downarrow$ \\ \midrule
Graph-based & 75.9 & 86.6 & 88.0 & 448.30 \\
w/o Tier 1 & 80.4 & 85.8 & 83.1 & 21.53 \\
w/o Tier 1, 2 & \textbf{85.9} & \textbf{89.7} & \textbf{88.8} & 56.05 \\
w/ Tier 1, 2 & 81.1 & 85.6 & 86.4 & \textbf{12.88} \\ \bottomrule
\end{tabular}
\end{table}

As shown in Table \ref{tab:ablation_hmo}, removing Tier 1 degrades accuracy to 83.1\% and increases latency to 21.53 min, proving that the first tier effectively filters noise to prevent redundant computation. While the exhaustive search w/o Tier 1, 2 achieves a peak accuracy of 88.8\%, it requires a prohibitive 56.05 min. In contrast, our full configuration w/ Tier 1, 2 reduces latency to 12.88 min, achieving a 77.0\% speedup by pruning the search space with only a marginal 2.4\% accuracy drop. This hierarchical orchestration significantly outperforms the Graph-based baseline which requires 448.30 min due to excessive structural traversal overhead.

\begin{table}[t]
\centering
\renewcommand{\arraystretch}{0.9}
\setlength{\tabcolsep}{1mm}
\caption{Efficiency comparison of memory update strategies.}
\label{tab:ablation_update}
\begin{tabular}{l|cccc}
\toprule
\textbf{Strategy} & \textbf{Recall} $\uparrow$ & \textbf{Acc.} $\uparrow$ & \textbf{Build (s)} $\downarrow$ & \textbf{Inf. (min)} $\downarrow$ \\ \midrule
MLLM-Scoring-only & 80.7 & 86.1 & 1.63 & 15.05 \\
Similarity-only & \textbf{83.3} & \textbf{86.7} & \textbf{0.15} & 87.58 \\
\textbf{HMO (Hybrid)} & 81.1 & 86.4 & 0.19 & \textbf{12.88} \\ \bottomrule
\end{tabular}
\end{table}

Table \ref{tab:ablation_update} shows MLLM-Scoring-only reaches 86.1\% accuracy but suffers 1.63s update time. While Similarity-only achieves 86.7\% accuracy, its 87.58 min inference time exceeds 56.05 min global search due to redundant 1+2+3 layer traversal. Our Hybrid approach achieves the fastest 12.88 min inference speed with a negligible 0.19s update time, striking the optimal efficiency-accuracy balance.

\subsection{Further Analysis}




\textbf{Evaluation on Large-scale Datasets}.
While the efficiency gains on LongMemEval-S are consistent, the advantages of the HMO architecture become more pronounced as the dataset scales. On the larger LongMemEval-M, a full Tier 3 global search requires 1453.21s to achieve a 68.2\% recall. In contrast, HMO reduces the total time to 195.75s while maintaining a competitive 63.0\% recall. This 86.5\% reduction in search time with marginal recall loss validates the scalability and effectiveness of our hierarchical orchestration in handling massive memory banks.

\section{Conclusion}

In this paper, we propose Hierarchical Memory Orchestration, a multi-tiered memory framework that reduces reliance on exhaustive search by architectural organization. By leveraging an evolving user persona to govern a three-tiered hierarchy, HMO prunes the retrieval space while maintaining high reasoning alignment and interaction fluidity. Our evaluations demonstrate that HMO achieves state-of-the-art performance across multiple benchmarks, significantly outperforming traditional flat and graph-based retrieval systems. Furthermore, we provide a robust software implementation integrated into the OpenClaw platform, enabling a user-centric ecosystem where memory evolution is directly coupled with individual behavioral patterns. This synergy between hierarchical storage and personalized orchestration proves that persistent agents can achieve real-time efficiency without compromising the fidelity of long-term historical awareness.

\bibliographystyle{ACM-Reference-Format}
\bibliography{sample-base}

\newpage
\appendix

\section{Prompt Design}

To ensure the precision of memory tiering and the objectivity of our experimental evaluations, we carefully designed specialized prompt templates for the MLLM-based evaluator $\mathcal{E}$ and the final reasoning judge. 

Table \ref{tab:initial_importance_prompt} details the prompt used for \textbf{Initial Importance Scoring}. This template directs the model to evaluate interaction segments across three critical dimensions: Behavioral Alignment, Reasoning Utility, and Contextual Persistence. By providing a structured 1--10 scoring rubric, we minimize the variance in importance initialization, ensuring that pivotal user traits are prioritized within the hierarchical tiers.

Furthermore, to maintain a rigorous evaluation standard across diverse benchmarks, we employ an \textbf{LLM-as-a-judge} protocol, as shown in Table \ref{tab:llm_judge_prompt} \cite{Tan_2025_RMM}. This prompt enforces a binary evaluation logic (Yes/No), requiring the model to verify if a generated response strictly satisfies the ground-truth requirements. The inclusion of few-shot examples and explicit constraints on partial answers ensures that our reported accuracy and recall metrics are both consistent and reproducible.

\begin{table*}[t]
\centering
\renewcommand{\arraystretch}{1.1}
\setlength{\tabcolsep}{5mm}
\caption{Detailed Performance Breakdown on the LoCoMo Benchmark (GPT-4o-mini). Performance is measured by F1 scores across Single-Hop, Multi-Hop, Temporal, and Open-Domain categories. The \textbf{\textcolor{blue}{best}} results are highlighted in blue bold, and the \textcolor{cyan}{second-best} results are marked in cyan.}
\label{table:locomo_detailed_appendix}
\begin{tabular}{l|cccc|c}
\toprule
\multirow{2}{*}{\textbf{Method}} & \multicolumn{5}{c}{\textbf{Answer Prediction (F1)}} \\
& \textbf{Single-Hop} & \textbf{Multi-Hop} & \textbf{Temporal} & \textbf{Open-Domain} & \textbf{Overall} \\ \midrule
A-Mem \cite{A-MEM} & 44.65 & 27.02 & \textcolor{cyan}{45.85} & 12.14 & 23.30 \\
MemoryBank \cite{A-MEM} & 41.04 & 26.65 & 25.52 & 9.15 & 31.42 \\
Mem0 \cite{Chhikara_2025_Mem0} & 38.72 & 28.64 & \textbf{\textcolor{blue}{48.93}} & 47.65 & 42.13 \\
MemoryOS \cite{Kang_2025_MemoryOS} & 35.27 & \textbf{\textcolor{blue}{41.15}} & 20.02 & 48.62 & 42.84 \\
MemVerse \cite{Liu_2025_Memverse} & \textcolor{cyan}{46.11} & 29.69 & 30.84 & \textcolor{cyan}{49.23} & \textcolor{cyan}{43.44} \\ \midrule
\textbf{HMO (Ours)} & \textbf{\textcolor{blue}{48.24}} & \textcolor{cyan}{32.15} & 35.66 & \textbf{\textcolor{blue}{51.08}} & \textbf{\textcolor{blue}{45.65}} \\ \bottomrule
\end{tabular}
\end{table*}

\begin{table*}[!t]
\centering
\caption{Prompt template for MLLM-based initial importance scoring.}
\label{tab:initial_importance_prompt}
\begin{tabular}{p{0.97\linewidth}}
\toprule
\textbf{Initial Importance Scoring Evaluator $\mathcal{E}$} \\
\midrule

\textbf{Role} \\
You are an expert Cognitive Memory Engine. Your task is to evaluate a new interaction record between a User and an AI Agent to determine its ``Initial Importance'' ($I_m$) for long-term retention. \\

\textbf{Context} \\
$\bullet$ User Persona ($P$): \{\{user\_persona\}\} \\
$\bullet$ Interaction Record ($m$): \{\{memory\_content\}\} \\

\textbf{Scoring Dimensions} \\
1. \textit{Behavioral Alignment \& Identity Mapping}: Evaluate if the record captures distinctive user traits, specialized preferences, or recurring behavioral patterns. Higher scores are assigned to information that defines "who the user is" (e.g., specific aesthetic tastes, dietary restrictions, or professional paradigms) rather than generic interactions. \\
2. \textit{Reasoning Utility \& Logical Dependency}: Assess if the memory serves as a foundational "ground truth" for future task execution. This includes explicit user constraints, finalized decisions, or milestone-related facts that, if forgotten, would lead to logical inconsistency or repetitive prompting in downstream reasoning. \\
3. \textit{Contextual Persistence \& Goal Projection}: Determine the memory's lifespan and its relevance to long-term objectives. Prioritize records that link the current session to ongoing projects, multi-step workflows, or latent intents that are likely to resurface in future contexts beyond the immediate dialogue window. \\

\textbf{Scoring Rubric ($I_m \in [1, 10]$)} \\
$\bullet$ 9--10 (Pivotal): Core identity facts, critical user constraints, or major life/project milestones. Essential for persona alignment. \\
$\bullet$ 7--8 (High Utility): Specific preferences, recurring habits, or detailed technical decisions that directly influence future interactions. \\
$\bullet$ 4--6 (Informative): Useful context but not critical. Standard information that might be relevant but isn't a ``behavioral trait.'' \\
$\bullet$ 1--3 (Transient): Generic greetings, filler talk, or repetitive common knowledge that provides no unique insight into the user. \\

\textbf{Output Format} \\
Provide only the raw integer score between 1 and 10. Do not provide any explanation. \\
Score: [Integer] \\

\bottomrule
\end{tabular}
\end{table*}

\begin{table*}[!t]
\centering
\caption{Prompt template for LLM-as-a-judge evaluation.}
\label{tab:llm_judge_prompt}
\begin{tabular}{p{0.97\linewidth}}
\toprule
\textbf{LLM-as-a-judge Evaluation Prompt} \\
\midrule

\textbf{Role} \\
You are an expert language model evaluator. I will provide you with a question, a ground-truth answer, and a model-generated response. Your task is to determine whether the response correctly answers the question by following these evaluation rules: \\

\textbf{Evaluation Rules} \\
$\bullet$ Answer \textit{Yes} if the response contains or directly matches the correct answer. \\
$\bullet$ Answer \textit{Yes} if the response includes all necessary intermediate steps leading to the correct answer. \\
$\bullet$ Answer \textit{No} if the response provides only a partial answer or omits essential information. \\
$\bullet$ Answer \textit{No} if the response does not sufficiently address the question. \\

\textbf{Examples} \\
\textit{Example 1: Correct Response} \\
$\bullet$ Question: What is the capital of France? \\
$\bullet$ Ground-truth Answer: Paris \\
$\bullet$ Response: The capital of France is Paris. \\
$\bullet$ Output: Yes \\

\textit{Example 2: Incorrect Response} \\
$\bullet$ Question: What is the capital of France? \\
$\bullet$ Ground-truth Answer: Paris \\
$\bullet$ Response: France is a country in Europe. \\
$\bullet$ Output: No \\

\textbf{Additional Instructions} \\
1. Apply the evaluation criteria consistently. \\
2. Base your decision strictly on the information in the response. \\
3. Avoid subjective interpretations and adhere to the provided examples. \\

\textbf{Input Format} \\
$\bullet$ Question: \{\} \\
$\bullet$ Ground-truth Answer: \{\} \\
$\bullet$ Response: \{\} \\
\textbf{Output:} \\

\bottomrule
\end{tabular}
\end{table*}

\section{Detailed Performance Metrics on the LoCoMo Benchmark}
\label{app:locomo_metrics}

As illustrated in Table \ref{table:locomo_detailed_appendix}, we provide a comprehensive breakdown of the F1 scores across diverse reasoning categories on the LoCoMo benchmark. Our proposed HMO framework achieves a state-of-the-art overall F1 score of 45.65, demonstrating superior performance in most categories compared to established baselines. 

Notably, HMO exhibits significant strengths in \textbf{Single-Hop} and \textbf{Open-Domain} scenarios, reaching 48.24 and 51.08 respectively. This performance indicates that the three-tiered hierarchical architecture effectively maintains high-fidelity memory retrieval without the typical noise interference found in flat memory systems. While specialized methods like MemoryOS or Mem0 may show localized advantages in Multi-Hop or Temporal reasoning, HMO provides the most balanced and robust performance across the entire spectrum of personalized interaction tasks.

\section{Memory Dynamics and Benchmark Constraints}
\label{sec:memory_limitations}

A fundamental discrepancy exists between the operational logic of HMO and the evaluation protocols of current benchmarks like LoCoMo and LongMemEval. These datasets typically adopt a decoupled evaluation paradigm where the interaction history is provided as a static corpus and questions are treated as isolated retrieval tasks. This structure fails to account for the temporal coherence and iterative evolution that characterize how users and agents engage in the real world.

HMO is specifically engineered for in-situ interaction where memory orchestration occurs concurrently with the dialogue flow. In a realistic deployment, each user query serves as a trigger to redistribute Tiers and refine the Persona in real time. This allows the system to effectively pre-fetch and prioritize knowledge for subsequent inquiries that are logically connected. Our framework thrives when the conversation is a continuous and evolving stream where the agent response is informed by a memory state that was just updated by the preceding exchange. 

To bridge this gap, we will continue to investigate whether more robust benchmarks exist that rely on interaction and can effectively simulate this co-evolutionary process. The current lack of standardized evaluation suites that measure how an agent prepares its internal memory structure for future and coherent follow-ups remains a significant challenge. Consequently, we aim to further explore the potential for proposing a novel evaluation framework centered on the user. By focusing on live interactions and the dependency between multiple turns, such a benchmark could provide a more rigorous environment to validate the efficiency of HMO hierarchical fluidity in truly personalized and real-world ecosystems.

\section{System Interface and Interaction Examples}
\label{sec:ui_demonstration}

\begin{figure*}[t]
\centering
\fbox{\includegraphics[width=1.00\textwidth]{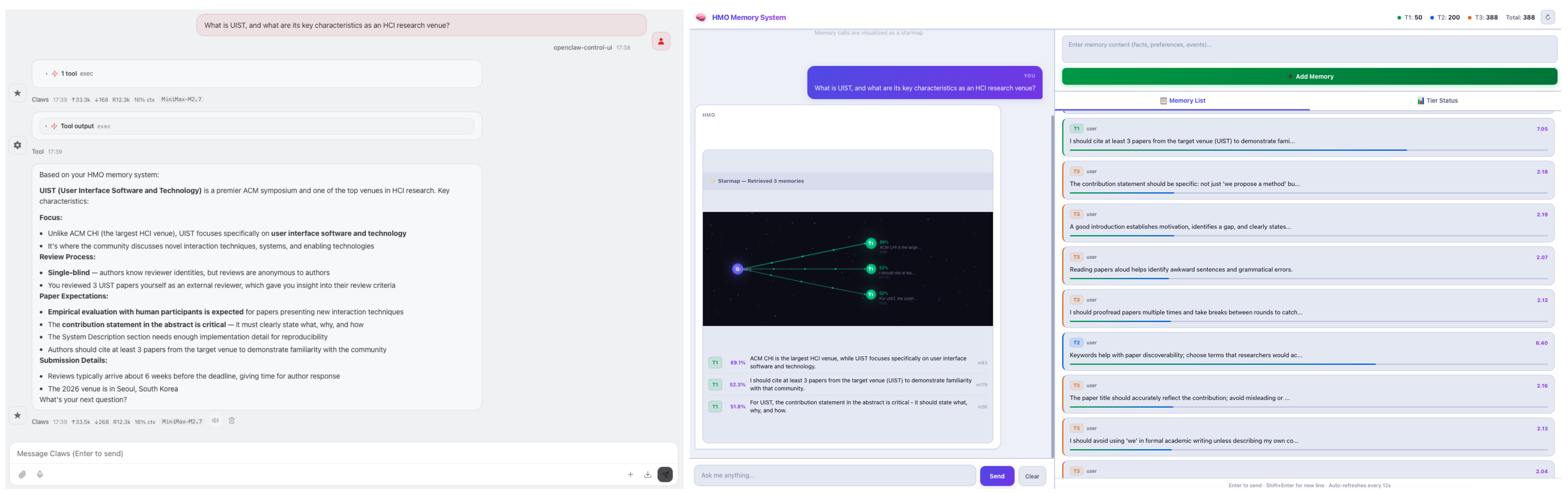}} 
\caption{System interface demonstration (example 1).}
\label{fig:ui_example1}
\end{figure*}

\begin{figure*}[b]
\centering
\fbox{\includegraphics[width=1.00\textwidth]{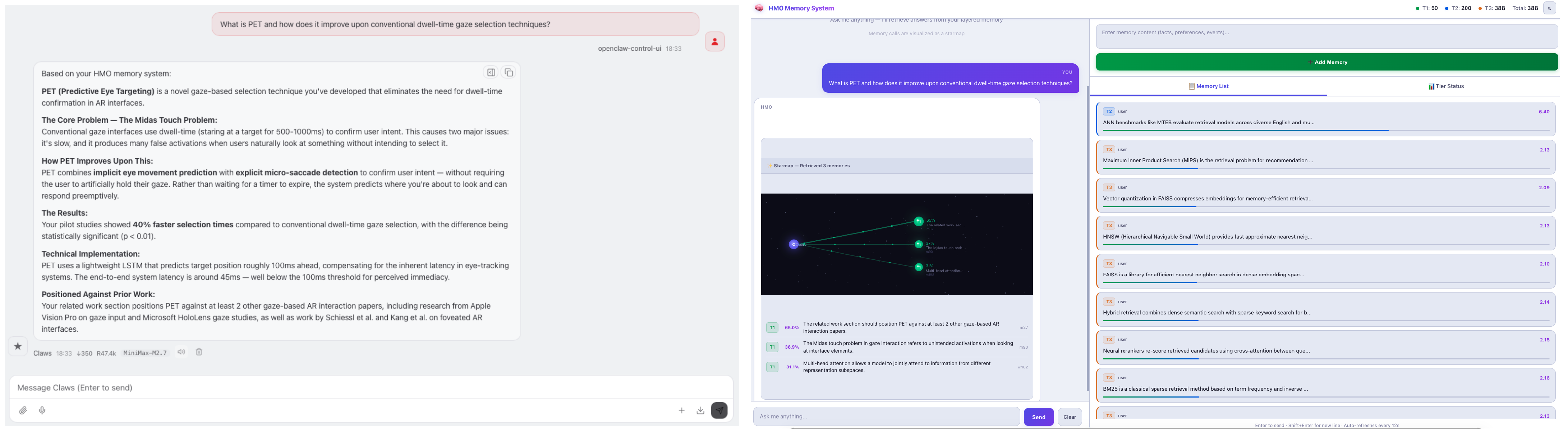}} 
\caption{System interface demonstration (example 2).}
\label{fig:ui_example2}
\end{figure*}

As illustrated in Figure \ref{fig:ui_example1} and Figure \ref{fig:ui_example2}, we provide additional examples of the system interface across various interaction scenarios. These cases demonstrate the consistent performance of HMO when dealing with different user types and dialogue topics.

To further evaluate the practical utility of our framework, we deployed HMO within a testing group consisting of 35 diverse users. The feedback from this pilot deployment was overwhelmingly positive, with participants noting that the agent maintained a more coherent and personalized understanding of their long-term preferences compared to previous systems. This real-world validation confirms that the hierarchical memory structure effectively aligns with user needs in active and continuous conversations.

\end{document}